\documentclass{article}

     \usepackage[preprint]{neurips_2022}

\usepackage[utf8]{inputenc} 
\usepackage[T1]{fontenc}    
\usepackage{hyperref}       
\usepackage{url}            
\usepackage{booktabs}       
\usepackage{amsfonts}       
\usepackage{nicefrac}       
\usepackage{microtype}      
\usepackage{xcolor}         
\usepackage{graphicx}
\usepackage{amsthm,amsmath,amssymb}
\usepackage{mathrsfs}
\usepackage{bbm}
\usepackage{dsfont}
\usepackage{amsmath, amsfonts, amssymb}
\usepackage{makecell}
\usepackage{mathrsfs}
\usepackage{stfloats}

\title{Learning from Positive and Unlabeled Data with Augmented Classes}

\author{%
	{Zhongnian Li$^1$, Liutao Yang$^2$, Zhongchen Ma$^3$, Tongfeng Sun$^1$,  Xinzheng Xu$^1$\thanks{
		Corresponding author} \, and Daoqiang Zhang$^2$} \\
	$^1$Department of Computer Science, China University of Ming and Technogy\\ $^2$Department of Computer Science, Nanjing University of Aeronautics and Astronautics \\ $^3$Department of Computer Science, Jiangsu University\\ 
	\texttt{zhongnianli@cumt.edu.cn, xxzheng@cumt.edu.cn} \\
}

\begin{document}

\maketitle

\begin{abstract}
Positive Unlabeled (PU) learning  aims to learn a binary classifier from only positive and unlabeled data, which is utilized in many real-world scenarios. However, existing PU learning algorithms cannot deal with the real-world challenge in an open and changing scenario, where examples from unobserved augmented classes may emerge in the testing phase. In this paper, we propose an unbiased risk estimator for PU learning with Augmented Classes (PUAC) by utilizing unlabeled data from the  augmented classes distribution, which can be easily collected in many real-world scenarios. Besides, we derive the estimation error bound for the proposed estimator, which provides a theoretical guarantee for its convergence to the optimal solution. Experiments on multiple realistic datasets demonstrate the effectiveness of proposed approach. 
\end{abstract}

\section{Introduction}

Learning from Positive and Unlabeled (PU)\cite{DBLP:journals/ml/BekkerD20, DBLP:conf/nips/NiuPSMS16, DBLP:conf/nips/PlessisNS14,DBLP:conf/pkdd/BekkerRD19, DBLP:conf/icml/PlessisNS15} data is a type of weakly supervised learning\cite{DBLP:conf/iclr/LuNMS19,DBLP:conf/nips/IshidaNS18,DBLP:conf/icml/BaoNS18,DBLP:journals/neco/ShimadaBSS21}, which has drawn considerable attention in many real-world scenarios. The goal of PU learning is to train a binary classifier by using only positive and unlabeled data without the assistance of negative label, which requires huge costs in some tasks. PU learning has many practical applications, such as, text classification, image annotation, time series categorization, bio-medicine analysis and so on.

Previous researches\cite{DBLP:conf/cvpr/GuoXHWSXT20, DBLP:journals/pami/SansoneNZ19,DBLP:conf/aaai/SakaiS19,DBLP:conf/iclr/KatoTH19} focus on handling  unlabeled data to  solve the PU learning problem. Specifically, a line of effective algorithms aim to extract possible negative examples from unlabeled data\cite{DBLP:conf/ijcai/LiL03,DBLP:journals/prl/MordeletV14}, and then  train a binary classifier. The performance of this category is heuristic and strongly influenced by selected  negative examples. Another category methods train PU learning classifiers by treating unlabeled data as noise negative examples. Thus, the classifier can be trained by utilizing the small weights for negative data\cite{DBLP:conf/nips/PlessisNS14, DBLP:conf/icml/PlessisNS15}, which heavily relies on the choice of weight for unlabeled data. 

However, it is noteworthy that exist studies on PU learning were in a stable scenario rather than non-stationary environment\cite{DBLP:journals/tkde/WeiYMWSZ21, DBLP:conf/icml/PhamRFA15,DBLP:conf/icml/GuoZJLZ20}, where some examples derived from unobserved classes in training phase might emerge in the testing data\cite{DBLP:conf/nips/Zhang0MZ20, DBLP:conf/aaai/DaYZ14}. In this paper, we focus on PU learning with Augmented Classes (PUAC). Augmented classes classification desire to make reliable prediction, which not only identifies augmented classes but also classifies the data form observed classes accurately in testing phase.  

\begin{figure}
	\centering
	\includegraphics[width=0.8\textwidth]{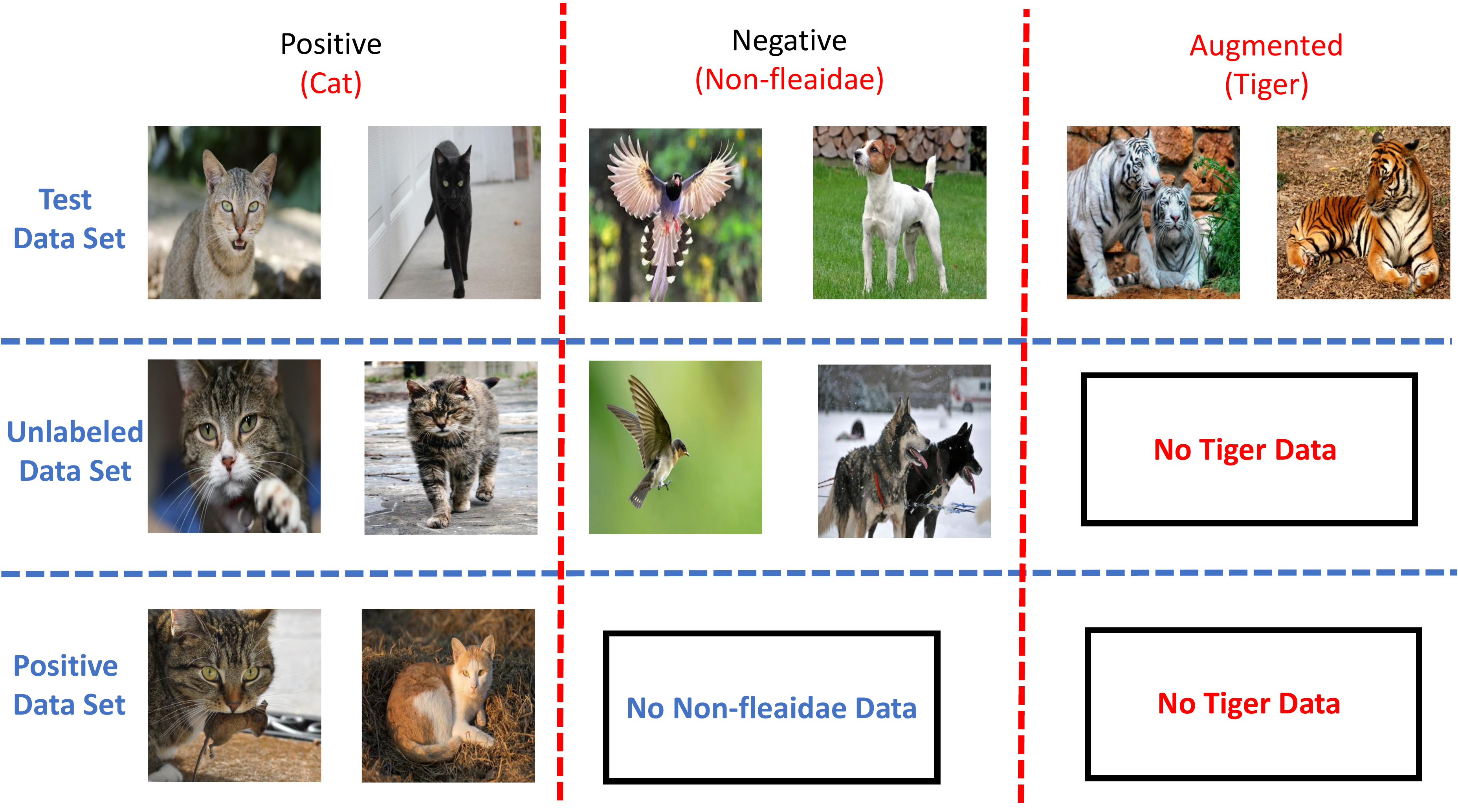}
	\caption{One example of Positive and Unlabeled data with  Augmented Classes. The augmented class (Tiger) only appear in testing dataset, which is not observed in positive and unlabeled data.}
	\label{motivation}
	\vspace{-1em}
\end{figure}

For example, in the task of felidae image annotation utilizing PU learning in the Internet, the user may only label cat  images as positive data and the absence label images as unlabeled data as shown in Fig.\ref{motivation}. In the traditional PU learning setting, unlabeled data only consist of cat and non-felidae images. However, since the environment is open and change, there may appear some images from augmented classes, such as tiger images. When an image of tiger comes, exist PU classifiers nearly predict it in the cat or non-felidae class, which will degrade performance of those classifiers. When facing some open and change scenarios in the real world, a mature PU learning algorithm need to work in PUAC setting.

In order to deal with the PUAC problem, we propose an unbiased risk estimator called UPUAC, short for \underline{U}nbiased risk estimator of \underline{P}ositive and \underline{U}nlabeled learning with \underline{A}ugmented \underline{C}lasses. UPUAC exploits unlabeled data form the  augmented classes distribution, which can be collected easily from many real-world scenarios. More concretely, we rewrite the risk into an equivalent expression by utilizing positive, unlabeled and  augmented classes distributions, which will lead to an unbiased risk estimators for PUAC. Besides, we provide a theoretical analysis of estimation error bound which certainly guarantees the estimator converges to the optimal solution. Experimentally, by comparing with existing state-of-the-art PU learning approach, our UPUAC achieves the best classification performance when the augmented classes emerges  on multiple realistic datasets. Beyond that, we test proposed UPUAC with some inaccurate class priors, which may be obtain by mixture proportion estimation\cite{DBLP:journals/ml/PlessisNS17,DBLP:conf/icml/RamaswamyST16}.    

\section{Preliminaries}
In this section, we describe  PUAC problem setting and review notations of risk rewrite briefly.

\subsection{Problem Setting of PUAC Learning}
In traditional PU learning, the learner collects two datasets $D_p = \{{ x}_i^p\}_{i=1}^{n_p}$ and $D_u = \{{x}_i^u\}_{i=1}^{n_u}$ sampled from positive distribution $P_p(x)$ and unlabeled distribution $P_u(x)$ respectively, where ${\bf x} \in \mathbb{R}^d$ is a $d$-dimensional feature and  $\mathcal{X}$ is the feature space. Let  $y_i \in \{p,n\}$ denotes the class label and $\mathcal{Y'} = \{p,n\}$ denotes the label space for traditional PU learning. In the PUAC problem, the learner requires to train a classifier by using the data form the  augmented classes distribution, where data from unobserved augmented classes might emerge. Since the number of augmented classes is unknown, the data generated form unobserved augmented classes will be predicted as a single class $\bf a$.  In our setup, an unlabeled dataset $D_a = \{{ x}_i^a\}_{i=1}^{n_a}$ sampled from the  augmented classes distribution $P_a(x)$, and Let $y_i \in \{p,n,a\}$ denotes the class label,  $\mathcal{Y} = \{p,n,a\}$ denotes the label space of PUAC.  Let  $p_p(x) = p(x|y = p)$, $p_n(x) = p(x|y = n)$,  $p_a(x) = p(x|y = a)$ denote the class-conditional densities for positive, unlabeled and  augmented classes distributions respectively, $\widetilde D = D_p \cup D_u \cup D_a$ denote the aggregated dataset, $P(x,y)$ denotes the distribution of $\widetilde D$,  $\pi_p = p(y=p)$, $\pi_n = p(y=n)$ and $\pi_a = p(y=a)$ denote the class prior probabilities for aggregated dataset.

{\bf Data generation process \;} Let $\theta_p^p$,$\theta_p^n$,$\theta_p^a$, $\theta_u^p$, $\theta_u^n$,$\theta_u^a$, $\theta_a^p$, $\theta_a^n$ and $\theta_a^a$ be  class priors for positive,  unlabeled and augmented classes distributions. The data collected for PUAC is assumed to be   i.i.d sample from the marginal densities as follows:
\begin{equation}
\begin{array}{l}
{P_{{p}}(x)} = \theta _p^p{p_p(x)}+ \theta _p^n{p_n(x)} + \theta _p^a{p_a(x)},\vspace{1ex}\\
{P_u(x)} = \theta _u^p{p_p(x)} + \theta _u^n{p_n(x)} + \theta _u^a{p_a(x)},\vspace{1ex}\\
{P_a(x)} = \theta _a^p{p_p(x)} + \theta _a^n{p_n(x)} + \theta _a^a{p_a(x)}
\end{array}
\end{equation}where $\theta _p^p = 1$, $\theta _p^n = \theta _p^a=0$, $\theta _u^p + \theta _u^n = 1$, $\theta _u^a = 0$ and $\theta _a^p + \theta _a^n + \theta _a^a = 1$. In the PUAC problem setting, the number of free class priors is three, which could be estimated by various methods that is similar to the traditional PU learning. 

Data generation process states that the distribution of PUAC data can be regarded as a mixture of positive, negative and augmented classes with some class priors. It is noteworthy that it does not matter that   augmented classes distribution equals to the testing distribution. It means that collecting data  from  augmented  classes distribution is easier than from the testing distribution, which is not a stable distribution during the collecting phase in the non-stationary environments. In the experiments, empirical results confirm the robustness for class distribution shifting in the testing distribution. 

\subsection{Risk Rewrite}
In supervised positive, negative and augmented classes classification, let $f:\mathcal{X} \to \mathcal{Y}$ denotes the decision function, i.e.,$f$ may be any multi-class classifier. Let $\mathnormal{l}:\mathcal{Y} \times \mathcal{Y} \to \mathbb{R}$ be the loss function, and given class priors $\pi_p = p(y=p)$, $\pi_n = p(y=n)$ and $\pi_a = p(y=a)$, the risk of $f$ is formulated as follows:
\begin{equation}
R_\mathnormal{l}(f) = {\mathbb{E}_{(x,y) \sim p(x,y)}}[\mathnormal{l}(f(x), y)] = \sum\limits_{i = p,n,a} {{\pi _i}{\mathbb{E}_{x \sim {p_i}}}[\mathnormal{l}(f(x), i)]}
\label{mcerm}
\end{equation} where $ p(x,y)$ denotes the distribution of supervised dataset, $p_i$ denotes the $i^{th}$ class-conditional density, i.e., $p_i(x)= P(x|y=i)$.


In the PUAC problem, since the negative and  augmented classes examples are unavailable,  we cannot  estimate ${\mathbb{E}_{x \sim {p_n}}}[\mathnormal{l}(f(x), n)]$ and ${\mathbb{E}_{x \sim {p_a}}}[\mathnormal{l}(f(x), a)]$, which means that Eq.\ref{mcerm} is unable to calculate for PUAC directly. Thus, we rewrite the risk  by using unlabeled and  augmented classes distribution to replace the supervised multi-class risk $R_l(f)$. The definition of risk rewrite for PUAC is shown as follows.

\textbf{Definition 1.} The risk $R_\mathnormal{l}(f)$ is rewritable for PUAC setting on the basis of three  marginal densities $P_{{p}}$, $P_{{u}}$ and $P_{{a}}$, and if and only if  there exist constants $\alpha_p$,$\alpha_n$,$\alpha_a$,$\beta_p$,$\beta_n$,$\beta_a$,$\gamma_p$,$\gamma_n$ and $\gamma_a$, such that for any $f$ it hold that 
\begin{equation}
R_\mathnormal{l}(f) =  {\mathbb{E}_{P_p}}[\mathnormal{ \widetilde l_p}(f(x))] + {\mathbb{E}_{P_u}}[\mathnormal{ \widetilde l_u}(f(x))] +{\mathbb{E}_{P_a}}[\mathnormal{ \widetilde l_a}(f(x))]
\label{acerm}
\end{equation} where $\widetilde l_p(f(x)) = \alpha_pl(f({{x}}),p) + \alpha_nl(f({{x}}),n) + \alpha_al(f({{x}}),a) $, $\widetilde l_u(f(x)) = \beta_pl(f({{x}}),p) + \beta_nl(f({\text{x}}),n) + \beta_al(f({\text{x}}),a) $ and $\widetilde l_a(f(x)) = \gamma_pl(f({\text{x}}),p) + \gamma_nl(f({\text{x}}),n) + \gamma_al(f({\text{x}}),a) $ are the corrected loss functions. 

If the risk is rewritable for PUAC setting, the PUAC risk can be expressed as expectation over $p_p(x)$, $p_n(x)$ and $p_a(x)$ separately, which builds a bridge to achieve an unbiased risk estimator.

\section{Learning from PUAC}
In this section, we first prove the risk is rewritable for PUAC setting. Then we describe the practical implementation for learning from positive and unlabeled learning with  augmented classes. In this paper, all the proofs can be found in Appendix.

\subsection{ Risk Rewrite for PUAC} 
Now, we attempt to express the supervised positive, negative and augmented classes risk $R_l(f)$ on  the basis of the three marginal densities given in Section 2.1. According to the Definition 1, if we find the constants $\alpha_p$,$\alpha_n$,$\alpha_a$,$\beta_p$,$\beta_n$,$\beta_a$,$\gamma_p$,$\gamma_n$ and $\gamma_a$, we could rewrite the PUAC risk. Fortunately, we show that those constants can be obtained by utilizing  class priors. An answer to find constants is given by the following Theorem 2. 

\textbf{Theorem 2.} Fix class priors  $\theta_p^p$, $\theta_u^p$, $\theta_u^n$, $\theta_a^p$, $\theta_a^n$ and $\theta_a^a$, then the supervised multi-classification $R_l(f)$ is rewritable, by letting $\alpha_p = \frac{\pi_p} {\theta_p^p} $,$\, \, \alpha_n = -\frac{\theta_u^p \pi_u}{\theta_p^p \theta_u^n}\,$,$\, \, \alpha_a =\frac{\theta_a^n \pi_a - \theta_a^p \theta_u^n \pi_a}{\theta_p^p \theta_u^n \theta_a^a}$,$\, \, \beta_p = 0$,$\, \,\beta_n = \frac{\pi_n}{\theta_u^n}$,$ \, \,\beta_a = -\frac{\theta_a^n \pi_a}{\theta_u^n \theta_a^a}$,$ \, \,\gamma_p = 0$,$ \, \,\gamma_n = 0$ and $ \, \,\gamma_a = \frac{\pi_a}{\theta_a^a}$.

 As the results of above theorem, we can express the PUAC risk using data collected from positive, unlabeled and  augmented classes distribution.  Let $\widetilde l_p(f(x))$, $\widetilde l_u(f(x)) $ and $\widetilde l_a(f(x))$ be 
\begin{align}
&\widetilde l_p(f(x)) = \frac{\pi_p} {\theta_p^p}l(f({\text{x}}),p) -\frac{\theta_u^p \pi_u}{\theta_p^p \theta_u^n}l(f({\text{x}}),n) + \frac{\theta_a^n \pi_a - \theta_a^p \theta_u^n \pi_a}{\theta_p^p \theta_u^n \theta_a^a}l(f({\text{x}}),a)\\
&\widetilde l_u(f(x)) =  \frac{\pi_n}{\theta_u^n}l(f({{x}}),n)  -\frac{\theta_a^n \pi_a}{\theta_u^n \theta_a^a}l(f({{x}}),a)\\
&\widetilde l_a(f(x)) =  \frac{\pi_a}{\theta_a^a}l(f({{x}}),a)
\end{align}

\textbf{Proposition 3.} The classification risk can be rewritten equivalently as 
\begin{equation}
\begin{split}
R_\mathnormal{PUAC,l}(f) &= {{\mathbb{E}_{x \sim {P_p}}}[\widetilde l_p(f(x))] + {\mathbb{E}_{x \sim {P_u}}}[\widetilde l_u(f(x))] + {\mathbb{E}_{x \sim {P_a}}}[\widetilde l_a(f(x))]} \\ &=
 \mathbb{E}_{x \sim {P_p}}\bigg[ \frac{\pi_p} {\theta_p^p}l(f({{x}}),p) -\frac{\theta_u^p \pi_u}{\theta_p^p \theta_u^n}l(f({{x}}),n) + \frac{\theta_a^n \pi_a - \theta_a^p \theta_u^n \pi_a}{\theta_p^p \theta_u^n \theta_a^a}l(f({{x}}),a)\bigg]  \\& \;\;\;\; + {\mathbb{E}_{x \sim {P_u}}}\big[\frac{\pi_n}{\theta_u^n}l(f({{x}}),n)  -\frac{\theta_a^n \pi_a}{\theta_u^n \theta_a^a}l(f({{x}}),a)\big] + {\mathbb{E}_{x \sim {P_a}}}[\frac{\pi_a}{\theta_a^a}l(f({{x}}),a)]
\end{split}
\label{mcerm_puac}
\end{equation} 

The Proposition 3 naturally leads to an unbiased estimator for PUAC. In the empirical minimization framework, Eq.\ref{mcerm_puac} is replaced with their empirical as follows:
\begin{equation}
\widehat R_\mathnormal{PUAC,l}(f) = {\frac{1}{{{n_p}}}\sum\limits_{i = 1}^{{n_p}}} \widetilde l_p(f(x)) + \frac{1}{{n_u}}\sum\limits_{i = 1}^{n_u} \widetilde l_u(f(x)) + \frac{1}{{n_a}}\sum\limits_{i = 1}^{n_a} \widetilde l_a(f(x))
\label{ecerm}
\end{equation} 
where the $n_p$, $n_u$ and $n_a$ denotes the number of examples collected form positive, unlabeled and augmented classes distributions respectively. In the rest of the paper,  the process of obtaining the empirical risk minimizer of Eq.\ref{ecerm}, i.e., ${\widehat f_{PUAC}} = \arg {\min _{f \in F}}{R_{PUAC,{{l}}}}(f)$ is named as PUAC learning. In this paper, we  propose  a ERM-based PUAC learning, and consequently,  ${\widehat f_{PUAC}} $ can be obtained by optimizing the Eq.\ref{ecerm}.
 
\textbf{Special case} Consider the Eq.\ref{ecerm} by specifying class priors. It is obvious that PUAC problem reduces to Positive and Unlabeled learning, if $\theta_u^n = 0 $ and $\theta_a^n = 0$ and the augmented classes data is seen as unlabeled examples. Then, we can train the classifier by using the positive and unlabeled learning. Besides, PUAC problem reduces to the standard Positive and Unlabeled learning, if $\theta_a^a = 0$. 


\subsection{Practical Implementation}
In this section, we investigate the practical implementation when the deep model is employed as classifier for PUAC.

\textbf{Loss function for PUAC problem} We investigate appropriate choice of the loss function $l$. Typically, the discrete loss is computationally hard, such as, $\mathds{1}[y\ne p]$, where $\mathds{1}(\bullet)$ denotes the indicator function. Thus, surrogate losses for multi-class learning with consistency properties\cite{DBLP:journals/jmlr/Zhang04a,DBLP:conf/icml/NarasimhanRS015} are used to replace discrete losses.   Some practical  examples of common multi-class loss function could be considered which satisfy the consistency proven in \cite{DBLP:journals/jmlr/Zhang04a}.  

One group of  the loss functions $\mathnormal{l}$ is the zero-one loss, i.e., $\mathnormal{l}_{0-1}(f(x),y)=\mathds{1}[f(x)\ne y]$. Then,  Eq.\ref{mcerm_puac} is  known as the classification error. In the multi-classification, the One-Versus-Rest (OVR) strategy with margin loss $\mathnormal{l_{mar}}$ has the guarantee that $R_\mathnormal{l_{mar}}(f_p, f_n, f_a)$ and $R_{l_{0-1}}(f_p, f_n, f_a)$ have the same minimizer over all decision functions \cite{DBLP:journals/jmlr/Zhang04a}. The $\mathnormal{l_{mar}}$ is formulated as,
\begin{equation}
l_{mar}(f_p(x), f_n(x), f_a(x), y) = \phi ({f_y}(x)) + \sum\limits_{\begin{subarray}{l} 
	i = p,n,a \\ 
	i \ne y 
	\end{subarray}}  {\phi ( - {f_i}(x))} 
\end{equation} where $f_i$ denotes the classifier for $i^{th}$ class, $\phi(z):\mathbb{R} \to[0, \infty)$ is a binary convex surrogate  loss, and $z=yf(x)$ denotes the margin. Many margin loss functions satisfy consistency properties for multi-class problem, such as square loss $\phi (z) = (1-z)^2$. Then, According to  the OVR strategy with margin loss, the loss  $\widetilde l_p(f_p(x), f_n(x), f_a(x)$  is formulated as  
\begin{equation}
\begin{split}
\widetilde l_p(f_p(x), f_n(x), f_a(x)) =& \frac{\pi_p} {\theta_p^p} [\phi ({f_p}(x))+\phi (-{f_n}(x)) + \phi (-{f_a}(x))] \\& -\frac{\theta_u^p \pi_u}{\theta_p^p \theta_u^n}[\phi (-{f_p}(x))+\phi ({f_n}(x)) + \phi (-{f_a}(x))] \\& + \frac{\theta_a^n \pi_a - \theta_a^p \theta_u^n \pi_a}{\theta_p^p \theta_u^n \theta_a^a}[\phi (-{f_p}(x))+\phi (-{f_n}(x)) + \phi ({f_a}(x))]
\end{split}
\label{mcerm_p_loss}
\end{equation} The similar formulation can be obtained for loss $\widetilde l_u(f(x))$ and $\widetilde l_a(f(x))$. After obtaining the binary classifier $f_i$, we can construct the multi-class predictor as $f:X \to Y$ with $f(x) = \arg {\max _{k \in \{ p,n,a\} }}{f_k}(x)$.

Another group of loss for multi-class problem is ordinal regression loss\cite{DBLP:conf/nips/FinocchiaroFW19,DBLP:journals/jmlr/RamaswamyA16,DBLP:journals/jmlr/PedregosaBG17}, e.g. using absolute distance:
\begin{equation}
{l_{ord}}(f_o(x),y) = \left| {f_o(x) - y} \right|\;\;\;\;\;\;\;\forall f_o(x),y \in \{ p,n,a\} 
\end{equation} where label $p=1$, $n=2$ and $a=3$, $f_o(x)$ denotes the multi-class classifier. Many surrogate losses satisfy consistency properties for ordinal regression loss in multi-class learning, such as absolute surrogate loss. Compared to zero-one surrogate loss with three dimension classifier, the prediction dimension of absolute surrogate is one, which can be employed to develop computationally efficient methods\cite{DBLP:conf/nips/FinocchiaroFW19}.  Let $\phi_y(f_o(x))$ denotes the absolute surrogate loss, the multi-class loss $l_{abs}$ can be formulated as  
\begin{equation}
l_{abs}(f_o(x),y) = \phi_y(f_o(x)) = \left| {f_o(x) - y} \right|\;\;\;\;\;\;\;\forall y \in \{ p,n,a\} ,f_o(x) \in R
\end{equation}We now give the risk minimization of Eq.\ref{mcerm_puac} as follows
\begin{equation}
\begin{split}
R_\mathnormal{PUAC,l_{abs}}(f_o) &=
\mathbb{E}_{x \sim {P_p}}\bigg[ \frac{\pi_p} {\theta_p^p}|f_o(x)-p| -\frac{\theta_u^p \pi_u}{\theta_p^p \theta_u^n}|f_o(x)-n| + \frac{\theta_a^n \pi_a - \theta_a^p \theta_u^n \pi_a}{\theta_p^p \theta_u^n \theta_a^a}|f_o(x)-a|\bigg]  \\& \;\;\;\; + {\mathbb{E}_{x \sim {P_u}}}\big[\frac{\pi_n}{\theta_u^n}|f_o(x)-n|  -\frac{\theta_a^n \pi_a}{\theta_u^n \theta_a^a}|f_o(x)-a|\big] + {\mathbb{E}_{x \sim {P_a}}}[\frac{\pi_a}{\theta_a^a}|f_o(x)-a|]
\end{split}
\label{mcerm_puac_abs}
\end{equation} 
Given the empirical loss of Eq.\ref{mcerm_puac_abs}, we can obtain the $\widehat f_{PUAC,o}$ by powerful stochastic optimization algorithm for deep models. After obtain the multi-class classifier, we can construct the predictor as  $f:X \to Y$ with $f(x) = \arg {\min _{k \in \{ p,n,a\} }}|{\widehat f_{PUAC,o}}(x)-k|$.

\textbf{Class Prior Estimation} It is noteworthy  that the optimization of  the proposed model requires estimating class priors, when the class priors were assumed to be unknown. In this section, we propose a class prior estimation algorithm from PU and augmented classes datasets.

Since we have samples collected from positive and unlabeled distributions, we can estimate $\theta _u^p$ and $\theta _a^p$ by mixture proportion estimation methods\cite{DBLP:conf/icml/RamaswamyST16} easily. For estimating $\theta _a^n$, we employ the kernel embedding proposed by, which use reproducing kernel Hilbert distance to estimate the mixture proportion. If $\theta _u^n$ and $\theta _u^p$ are estimated, class-conditional distribution can be expressed as ${p_n}(x) = \frac{{{P_u}(x) - \theta _u^p{P_p}(x)}}{{\theta _u^n}}$. Then, the kernel mapping distribution can be obtained from positive and unlabeled datasets. Since we have some examples collected from augmented classes distribution, the kernel mean distance can be computed between $p_n(x)$ and augmented classes distribution $P_a$ from those datasets. Thus, class prior $\theta _a^n$ can be estimated for PUAC problem.

\section{Theoretical Analysis}
In this section, we first study consistency property of proposed PUAC risk $R_\mathnormal{PUAC,l_{mar}}$. Then, we provide the theoretical analysis of error bound. 

Now, we show the Bayes classifiers for OVR strategy with zero-one loss $l_{0-1}$, and show that the PUAC risk $R_\mathnormal{PUAC,l_{mar}}$ is consistent with the supervised multi-class risk $R_\mathnormal{l_{0-1}}$. 

\textbf{Definition 4.} (Bayes classifiers for zero-one loss). Let $f^*:X \to Y$ be a classifier, $P(y = c\left| x \right.)$ denotes the conditional density. Then, the optimal classifier is obtained by minimizing the classification error, which often was referred to  the  Bayes classifiers for zero-one loss given by
\begin{equation}
{f^ * }(x) = \arg {\max _{c \in \{ p,n,a\} }}P(y = c\left| x \right.)
\end{equation}

The following theorem states that by minimizing the risk of $R_\mathnormal{PUAC,l_{mar}}$ , we can obtain the Bayes classifier for zero-one loss.

\textbf{Theorem 5.} Let $f_p$, $f_p$ and $f_a$ denotes binary classifiers for positive, negative and augmented classes, and $f(x) = \arg {\max _{k \in \{ p,n,a\} }}{f_k}(x)$, the surrogate loss $\phi(z)$ is convex, bounded below, differentiable, and $\phi(z) < \phi(-z)$ when $z> 0 $, then for any $\epsilon_1 > 0$, there exists $\epsilon_2 > 0$ such that 
\begin{equation}
R_{PUAC,{l_{mar}}}({f_p},{f_n},{f_a}) \leqslant R_{PUAC,{l_{mar}}}^ *  + {\varepsilon _2} \Longrightarrow {R_{0 - 1}}(f) \leqslant {R_{0-1}^ * } + {\varepsilon _1}
\end{equation} where $R_{PUAC,{l_{mar}}}^ * = min_{{f_p},{f_n},{f_a}}R_{PUAC,{l_{mar}}}({f_p},{f_n},{f_a})$ and $R_{0-1}^ * = min_{f}R_{0-1}(f)= R_{0-1}(f^*)$ denotes the Bayes error for multi-class distribution. 

Theorem 5 analysis the consistency property for proposed method, which means that we can obtain classifier achieving Bayes rule. According to Definition 4, it is obvious that the Bayes classifiers will achieve optimal classification error under class probabilities shift as usual in the testing distribution. By minimizing the PUAC risk $R_{PUAC,{l_{mar}}}$, we can get well-behaved classifiers the same as learning with supervised multi-class data. 

Now, we analyze the generalization error bounds for the proposed approach implemented by deep neural networks using OVR strategy. Let $ \mathbf f = ({f_p},{f_n},{f_a})$ denotes classification vector function in the deep network hypothesis set $\mathcal{F}$. Assume there is $C_\phi>0$, such that  $su{p_{z}}\phi(z)\leqslant C_{\phi}/3$, $K_a = max(\frac{\pi_p} {\theta_p^p},\frac{\theta_u^p \pi_u}{\theta_p^p \theta_u^n},|\frac{\theta_a^n \pi_a - \theta_a^p \theta_u^n \pi_a}{\theta_p^p \theta_u^n \theta_a^a}|)$, $K_u = max(\frac{\pi_n}{\theta_u^n},\frac{\theta_a^n \pi_a}{\theta_u^n \theta_a^a})$ and $K_a = \frac{\pi_a}{\theta_a^a}$. Let $L_\phi$ be the Lipschitz constant of $\phi$, we can establish the following lemma.

\textbf{Lemma 6.} For any $\delta>0$, with the probability at least $1-\delta/2$, 
 \begin{equation*}
{\sup _{\mathbf f \in \mathcal{F}}}\left| {{R_p}(\mathbf f) -  {\widehat{R}_p}  (\mathbf f)} \right| \leqslant 6{K_p}{L_\phi }{\mathfrak{R}_{{n_p}}}(\mathcal{F}) + 3{K_p}{C_\phi }\sqrt {\frac{{\ln (4/\delta )}}{{2{n_p}}}} 
 \end{equation*}
 \begin{equation*}
{\sup _{\mathbf f \in \mathcal{F}}}\left| {{R_u}(\mathbf f) -  {\widehat{R}_u}  (\mathbf f)} \right| \leqslant 4{K_u}{L_\phi }{\mathfrak{R}_{{n_u}}}(\mathcal{F}) + 2{K_u}{C_\phi }\sqrt {\frac{{\ln (4/\delta )}}{{2{n_u}}}} 
\end{equation*}
 \begin{equation*}
{\sup _{\mathbf f \in \mathcal{F}}}\left| {{R_a}(\mathbf f) -  {\widehat{R}_a}  (\mathbf f)} \right| \leqslant 2{K_a}{L_\phi }{\mathfrak{R}_{{n_a}}}(\mathcal{F}) + {K_a}{C_\phi }\sqrt {\frac{{\ln (4/\delta )}}{{2{n_a}}}} 
\end{equation*}where $R_p(\mathbf f) = \mathbb{E}_{x \sim {P_p}}[\widetilde l_p(\mathbf f(x))]$, $R_u(\mathbf f) = \mathbb{E}_{x \sim {P_u}}[\widetilde l_u(\mathbf f(x))]$ and $R_a(\mathbf f) = \mathbb{E}_{x \sim {P_a}}[\widetilde l_a(\mathbf f(x))]$, ${\widehat{R}_p}  (\mathbf f)$,  ${\widehat{R}_u}  (\mathbf f)$ and  ${\widehat{R}_a}  (\mathbf f)$ denote the empirical risk estimator to  $R_p(\mathbf f)$, $R_u(\mathbf f)$ and $R_a(\mathbf f)$ respectively, $\mathfrak{R}_{{n_p}}(\mathcal{F})$, $\mathfrak{R}_{{n_u}}(\mathcal{F})$ and $\mathfrak{R}_{{n_a}}(\mathcal{F})$ are the Rademacher complexities\cite{mohri2018foundations} of $\mathcal{F}$ for the sampling of size $n_p$ from $P_p(x)$, the sampling of size $n_u$ from $P_u(x)$ and the sampling of size $n_a$ from $P_a(x)$.

Based on the Lemma 6, we can obtain the estimation error bound as follows.

\textbf{Theorem 7.} For any $\delta>0$, with the probability at least $1-\delta/2$,
\begin{equation*}
\begin{split}
R_{l_{mar}}({\hat {\mathbf f}_{puac})} - \mathop{\rm {min}} _{{\mathbf{f}} \in \mathcal{F}}R_{l_{mar}}(\mathbf f) \leqslant & 12{K_p}{L_\phi }{\mathfrak{R}_{{n_p}}}(\mathcal{F}) + 8{K_u}{L_\phi }{\mathfrak{R}_{{n_u}}}(\mathcal{F}) + 4{K_a}{L_\phi }{\mathfrak{R}_{{n_a}}}(\mathcal{F}) \\&+ 3{K_p}{C_\phi }\sqrt {\frac{{\ln (4/\delta )}}{{2{n_p}}}}+ 2{K_u}{C_\phi }\sqrt {\frac{{\ln (4/\delta )}}{{2{n_u}}}}+ {K_a}{C_\phi }\sqrt {\frac{{\ln (4/\delta )}}{{2{n_a}}}}
\end{split}
\end{equation*} where $\hat {\mathbf f}_{puac}$ is trained by minimizing the PUAC risk $R_{PUAC,{l_{mar}}}$

Lemma 6 and Theorem 7 show that, with a growing number of positive, unlabeled and augmented classes data, the estimation error of the trained classifiers decreases, which means that the proposed method is consistent. When deep network hypothesis set $\mathcal{F}$ is fixed and  $\mathfrak{R}_{{n}}(\mathcal{F}) \leqslant C_{\mathcal{F}}/\sqrt{n}$, we  have $\mathfrak{R}_{{n_p}}(\mathcal{F}) = \mathcal{O}(1/\sqrt{n_p})$, $\mathfrak{R}_{{n_u}}(\mathcal{F}) = \mathcal{O}(1/\sqrt{n_u})$ and $\mathfrak{R}_{{n_a}}(\mathcal{F}) = \mathcal{O}(1/\sqrt{n_a})$, then 
\begin{equation*}
{n_p},{n_u},{n_a} \to \infty  \Longrightarrow R_{l_{mar}}({\hat {\mathbf f}_{puac})} - \mathop{\rm {min}} _{{\mathbf{f}} \in \mathcal{F}}R_{l_{mar}}(\mathbf f)  \to 0
\end{equation*}

Lemma 6 and Theorem 7 theoretically justify the effective of proposed method in exploiting augmented classes data.

\section{Experiments}
In this section, we experimentally analyze the proposed approach  from three aspects: 1) Classification accuracy for each class and identification accuracy for augmented classes  comparing with state-of-the-art methods; 2) Robustness for inaccurate training class priors; 3) Accuracy for class distribution shifting in the testing distribution. 

\subsection{Performance Comparison}

\textbf{Datasets:} In this section, we conduct experiments on five datasets, i.e., MNIST, Fashion-MNIST, Kuzushiji-MNIST, SVNH and CIFAR-10. The MNIST, Fashion-MNIST and  Kuzushiji-MNIST datasets consists of 70000 examples which originally have 10 classes. The SVHN dataset consists of 73257 examples with 10 classes. The CIFAR-10 dataset consists of 60000 examples associated with a label form 10 classes. We constructed the positive, unlabeled and augmented classes datasets as follows: we first select three classes form original datasets  as positive, negative and augmented classes datasets respectively. Then, we randomly select examples from positive dataset as positive class dataset, examples form positive and negative datasets as unlabeled dataset,  and  examples from positive , negative and augmented classes as augmented dataset. In the constructed procedure,  each example is  selected into only one dataset.  For performance comparison, we use the original testing datasets as the testing datasets.

\begin{table*}
	\centering
	\caption{Classification accuracy of each algorithm on benchmark datasets, with varying classes and the number of examples. P = 1, N=3, A=5 means that classes ${1}, 3,5 $ are taken as positive, negative, and augmented classes respectively. $\# $ PU and $\# $ AC denote the number of training examples in PU and augmented classes datasets. We report the mean and standard deviation of results over 5 trials. The best method is shown in bold (under 5$\%$ t-test). }
	{\resizebox{0.92\textwidth}{21mm}{
			\begin{tabular}{lc|cc|ccc cc }
				
				\toprule
				Dataset&P, N, A& $\# $PU&$\# $AC&UPU&NNPU&MPU&AREA&UPUAC\\
				\cmidrule{1-9} 
				&1,3,5&7591&10703&37.77$\pm$4.71&40.38$\pm$1.42&73.65$\pm$0.58&57.27$\pm$4.22&\textbf{98.90$\pm$0.24}\\ 
				&1,3,5&11320&6680&46.41$\pm$8.67&41.64$\pm$5.15&90.39$\pm$1.57&86.39$\pm$0.33&\textbf{98.70$\pm$0.11}\\
				MNIST&5,3,1&10625&7669&42.90$\pm$1.84&42.33$\pm$2.98&90.14$\pm$1.12&83.16$\pm$6.09&\textbf{99.47$\pm$0.08}\\ 
				&3,5,1\&2&10102&14150&25.24$\pm$2.06&27.20$\pm$2.30&89.66$\pm$0.33&77.86$\pm$1.68&\textbf{99.42$\pm$0.13}\\
				&9,8,7\&6&10313&13670&39.13$\pm$3.78&42.04$\pm$1.96&91.10$\pm$0.41&72.68$\pm$5.37&\textbf{98.39$\pm$0.21}\\
				\cmidrule{1-9} 
				&1,3,5&10335&7665&39.98$\pm$3.71&34.14$\pm$11.76&94.51$\pm$0.10&87.70$\pm$0.81&\textbf{98.60$\pm$0.06}\\
				Fashion&2,5,8&8650&9350&33.46$\pm$1.30&37.00$\pm$6.90&89.43$\pm$0.47&65.01$\pm$3.00&\textbf{98.80$\pm$0.08}\\ 
				&9,8,7&8658&9342&29.22$\pm$1.05&27.03$\pm$3.05&86.45$\pm$1.37&65.22$\pm$1.52&\textbf{97.30$\pm$0.11}\\ 
				\cmidrule{1-9} 
				&1,3,5&8674&9326&33.74$\pm$1.05&31.52$\pm$0.51&74.72$\pm$0.55&49.49$\pm$1.27&\textbf{94.31$\pm$0.52}\\
				Kuzushiji&2,4,6&10320&7680&31.44$\pm$3.33&31.05$\pm$2.56&77.36$\pm$0.32&53.97$\pm$0.19&\textbf{93.87$\pm$0.50}\\ 
				&9,8,6&9987&8013&36.12$\pm$6.92&35.18$\pm$3.66&81.34$\pm$0.71&57.90$\pm$1.47&\textbf{96.15$\pm$0.33}\\ 
				\bottomrule
	\end{tabular}}}
	\label{tab_acc_mnist}
		\vspace{-1em}	
\end{table*}
\begin{table*}
	\centering
	\caption{Identification accuracy of augmented classes on benchmark datasets, with varying classes and the number of examples. We report the mean and standard deviation of results over 5 trials. P = 1, N=3, A=5 means that classes ${1}, 3,5 $ are taken as positive, negative, and augmented classes respectively.  $\# $ train AC denotes the number of training examples in augmented classes datasets. The best method is shown in bold (under 5$\%$ t-test). }
	{\resizebox{0.99\textwidth}{10mm}{
			\begin{tabular}{l|ccc|ccc| ccc }
				
				\toprule
				Dataset&&MNIST&&&Fashion&&&Kuzushiji&\\
				\cmidrule{1-10} 
				P, N, A&5, 3, 1&1, 3, 5&9, 8, 7&1, 3, 5&2, 5, 8&9, 8, 7&1, 3, 5&3, 5, 1&9, 8, 7\\ 
				$\#$train AC &9955&10703&11173&7665&9350&9342&9326&8304&8013\\
				MPU&84.68$\pm$1.33&93.55$\pm$1.52&75.73$\pm$6.93&91.90$\pm$1.34&94.42$\pm$0.33&82.72$\pm$3.37&75.42$\pm$5.31&68.78$\pm$4.97&72.78$\pm$2.52\\ 
				AREA&59.05$\pm$1.79&72.77$\pm$4.50&54.01$\pm$3.83&88.22$\pm$2.27&52.98$\pm$10.21&49.62$\pm$9.59&44.62$\pm$8.49&39.70$\pm$10.12&47.70$\pm$5.36\\
				UPUAC&\textbf{98.22$\pm$ 0.48}&\textbf{98.89$\pm$0.15}&\textbf{97.28$\pm$0.55}&\textbf{96.98$\pm$0.31}&\textbf{98.86$\pm$0.49}&\textbf{97.74$\pm$1.00}&\textbf{96.36$\pm$1.00}&\textbf{97.14$\pm$0.27}&\textbf{92.92$\pm$1.10}\\
				\bottomrule
	\end{tabular}}}
	%
	\label{tab_tpr_mnist}	
		\vspace{-1em}
\end{table*}
\textbf{Common Setup:} We conduct experiments using OVR strategy implemented by margin square loss $\phi (z) = (1-z)^2$. As a classifier, we also used  neural network with 4 convolutional layers and 2 fully-connected layers for CIFAR-10,  neural network with 3 convolutional layers and 2 fully-connected layers for SVHN and 2 convolutional layers and 2 fully-connected layers for all the MNIST datasets. We used Adadelta\cite{DBLP:journals/corr/abs-1212-5701} for optimization and squared loss for experiments.

There are four contenders, consisting of binary PU learning approaches and multi-class PU learning approaches. We also report the classification accuracy on each class and identification accuracy for augmented classes. The details about the compared methods are described below.

\textbf{Binary PU Learning:} UPU\cite{DBLP:conf/icml/PlessisNS15} and NNPU\cite{DBLP:conf/nips/KiryoNPS17} are state-of-the-art binary PU learning approaches. We use the logistic loss for UPU and the sigmoid loss for NNPU. To compare with two binary PU learning approaches fairly, we merge the unlabeled and augmented classes datasets as new unlabeled dataset for binary classifier. Then for the testing phase, the predicted class of classifier is given  as $argmax\{f_p,f_n\}$ and the accuracy of classifier is calculated the same as multi-classes classification.  

\textbf{Multi-class PU Learning:} MPU\cite{DBLP:conf/ijcai/XuX0T17} and AREA\cite{DBLP:conf/icdm/ShuLY020} are state-of-the-art multi-class PU learning approaches. For comparing with two approaches fairly, the positive and unlabeled datasets were treads as two positive classes datasets and augmented classes dataset was treats as unlabeled dataset. In the testing phase, the predicted class is given as $argmax\{f_p,f_n, f_a\}$.

\begin{figure}
	\vspace{-0em}
	\centering
	\includegraphics[width=0.95\textwidth]{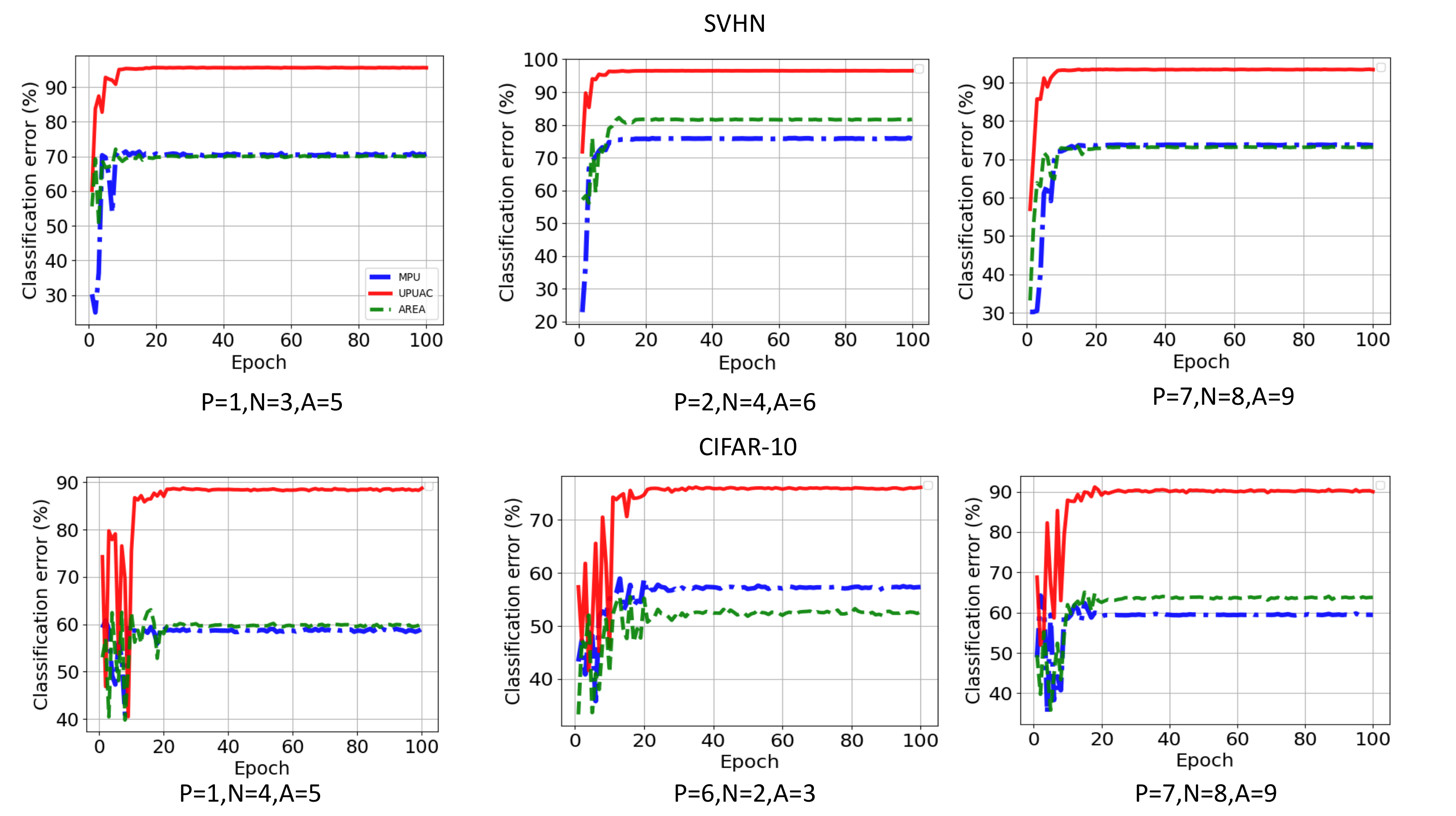}
	\caption{Illustrations classification accuracy of two datasets in experiments with various positive (P), negative (N) and augmented classes (A), datasets. }
	\label{results}
	\vspace{-0em}
\end{figure}

\begin{table*}
	\vspace{-1em}
	\centering
	\caption{Classification accuracy for inaccurate training class priors on MNIST dataset. P = 1, N=3, A=5 means that classes ${1}, 3,5 $ are taken as positive, negative, and augmented classes respectively. $\# $ PU and $\# $ AC denote the number of training examples in PU and augmented classes datasets. We report mean and standard deviation over 3 trials for varying degrees of inaccuracies. }
	{\resizebox{0.95\textwidth}{12mm}{
			\begin{tabular}{lc|cc|ccc cc }
				
				\toprule
				Dataset&P, N, A& $\# $PU&$\# $AC&\makecell[c]{$\eta_u^p = 0.8$ \\ $\eta_a^p=0.8$\\ $\eta_a^n=0.8$} &\makecell[c]{$\eta_u^p = 0.9$ \\ $\eta_a^p=0.9$\\ $\eta_a^n=0.9$}&\makecell[c]{$\eta_u^p = 1$ \\ $\eta_a^p=1$\\ $\eta_a^n=1$}&\makecell[c]{$\eta_u^p = 1.1$ \\ $\eta_a^p=1.1$\\ $\eta_a^n=1.1$}&\makecell[c]{$\eta_u^p = 1.2$ \\ $\eta_a^p=1.2$\\ $\eta_a^n=1.2$}\\
				\cmidrule{1-9} 
				&1,3,5&10752&7542&99.44$\pm$0.07&99.23$\pm$0.17&99.20$\pm$0.13&99.30$\pm$0.14&99.35$\pm$0.08\\ 
				MNIST&5,3,1&9659&8635&99.16$\pm$0.04&99.27$\pm$0.11&99.29$\pm$0.08&99.22$\pm$0.10&99.25$\pm$0.04\\ 
				&2,4,6&9771&7947&99.09$\pm$0.05&99.06$\pm$0.16&99.25$\pm$0.07&99.00$\pm$0.30&99.17$\pm$0.14\\
				\bottomrule
	\end{tabular}}}
	{\resizebox{0.95\textwidth}{12mm}{
			\begin{tabular}{lc|cc|ccc cc }
				
				\toprule
				Dataset&P, N, A& $\# $PU&$\# $AC&\makecell[c]{$\eta_u^p = 0.8$ \\ $\eta_a^p=1$\\ $\eta_a^n=1.2$} &\makecell[c]{$\eta_u^p = 0.9$ \\ $\eta_a^p=1$\\ $\eta_a^n=1.1$}&\makecell[c]{$\eta_u^p = 1$ \\ $\eta_a^p=0.8$\\ $\eta_a^n=1.2$}&\makecell[c]{$\eta_u^p = 1$ \\ $\eta_a^p=1.1$\\ $\eta_a^n=0.9$}&\makecell[c]{$\eta_u^p = 1.2$ \\ $\eta_a^p=0.8$\\ $\eta_a^n=1$}\\
				\cmidrule{1-9} 
				&1,3,5&10752&7542&99.20$\pm$0.20&99.26$\pm$0.24&99.38$\pm$0.19&99.24$\pm$0.06&99.30$\pm$0.15\\ 
				MNIST&5,3,1&9659&8635&99.27$\pm$0.07&99.30$\pm$0.14&99.20$\pm$0.07&99.26$\pm$0.02&99.17$\pm$0.17\\ 
				&2,4,6&9771&7947&98.94$\pm$0.38&99.13$\pm$0.14&99.08$\pm$0.19&99.13$\pm$0.20&99.10$\pm$0.14\\
				\bottomrule
	\end{tabular}}}
	\label{tab_acc_rou_mnist}
		\vspace{-0.5em}	
\end{table*}
Table.\ref{tab_acc_mnist} reports the classification accuracy of each algorithm on three datasets. It is obvious that the proposed method achieves the best result of all the binary PU learning and multi-class PU learning methods. Note that the performance of multi-class PU learning approaches MPU and AREA is better than binary  PU learning approaches, since the positive and augmented classes can be learned for multi-class classifiers. The noise only exists in negative examples for learning multi-class classifier. The similar trends are shown on identification accuracy. Table.\ref{tab_tpr_mnist} reports the identification  accuracy as well as the standard deviation of each algorithm. Fig.\ref{results} reports the experimental results on SVNH and CIFAR-10. We observer that the classification accuracy of MPU and AREA is very similar. Moreover, UPUAC is significantly better than multi-class positive and unlabeled learning methods.
\begin{table*}
	\vspace{-0em}
	\centering
	\caption{Classification accuracy of proposed method for  class distribution shifting on MNIST dataset. P = 3, N=4, A=5 means that classes ${3}, 4,5 $ are taken as positive, negative, and augmented classes respectively. $\# $ PU and $\# $ AC denote the number of training examples in PU and augmented classes datasets. We report  mean and standard deviation over 3 trials for varying degrees of shift. }
	{\resizebox{0.95\textwidth}{12mm}{
			\begin{tabular}{lc|cc|ccc cc }
				
				\toprule
				Dataset&P, N, A& $\# $PU&$\# $AC&\makecell[c]{$\eta_p = 0.8$ \\ $\eta_n=1$\\ $\eta_a=1.2$} &\makecell[c]{$\eta_p = 0.9$ \\ $\eta_n=1$\\ $\eta_a=1.1$}&\makecell[c]{$\eta_p = 1$ \\ $\eta_n=1$\\ $\eta_a=1$}&\makecell[c]{$\eta_p = 1.2$ \\ $\eta_n=1$\\ $\eta_a=0.8$}&\makecell[c]{$\eta_p = 1.1$ \\ $\eta_n=1$\\ $\eta_a=0.9$}\\
				\cmidrule{1-9} 
				&7,8,9&10109&7956&98.16$\pm$0.33&97.80$\pm$0.12&98.10$\pm$0.20&98.02$\pm$0.25&98.10$\pm$0.48\\ 
				MNIST&9,8,7&9826&8239&98.95$\pm$0.11&98.86$\pm$0.16&98.95$\pm$0.15&98.63$\pm$0.05&98.67$\pm$0.25\\ 
				&3,4,5&9955&7439&99.56$\pm$0.12&99.70$\pm$0.09&99.54$\pm$0.25&99.56$\pm$0.20&99.61$\pm$0.08\\
				\bottomrule
	\end{tabular}}}
	{\resizebox{0.95\textwidth}{12mm}{
			\begin{tabular}{lc|cc|ccc cc }
				
				\toprule
				Dataset&P, N, A& $\# $PU&$\# $AC&\makecell[c]{$\eta_p = 0.8$ \\ $\eta_n=1.1$\\ $\eta_a=1.2$} &\makecell[c]{$\eta_p = 0.9$ \\ $\eta_n=1.1$\\ $\eta_a=1.1$}&\makecell[c]{$\eta_p = 0.9$ \\ $\eta_n=0.8$\\ $\eta_a=1.2$}&\makecell[c]{$\eta_p = 0.9$ \\ $\eta_n=1.2$\\ $\eta_a=0.9$}&\makecell[c]{$\eta_p = 1.2$ \\ $\eta_n=1.2$\\ $\eta_a=0.9$}\\
				\cmidrule{1-9} 
				&7,8,9&10109&7956&97.79$\pm$0.33&97.73$\pm$0.25&97.77$\pm$0.17&97.87$\pm$0.03&97.83$\pm$0.50\\ 
				MNIST&9,8,7&9826&8239&98.87$\pm$0.17&98.80$\pm$0.19&98.82$\pm$0.17&98.79$\pm$0.20&98.66$\pm$0.07\\ 
				&3,4,5&9955&7439&99.63$\pm$0.07&99.73$\pm$0.03&99.76$\pm$0.03&99.68$\pm$0.11&99.45$\pm$0.03\\
				\bottomrule
	\end{tabular}}}
	\label{tab_acc_shift_mnist}	
	\vspace{-1em}
\end{table*} 
\subsection{Robustness for Inaccurate Training Class Priors \label{ritc}}
In above section, we have assumed that the class priors are accessible at the time of training. Here, we study the robustness for inaccurate training class priors, which can be estimated with mixture proportion estimation. Without loss of generality, we conduct the experiments on varying degrees of inaccuracies for class priors in the training phase. Let $\eta_u^p$,  $\eta_a^p$ and $\eta_a^n$ be real number around 1, $\vartheta _u^p = \eta_u^p \theta _u^p$,  $\vartheta _a^p = \eta_a^p \theta _a^p$ and  $\vartheta _a^n = \eta_a^n \theta _a^n$ be perturbed class priors. In this section, we draw data by using  $\theta _u^p$, $\theta _a^p$ and $\theta _a^n$ but train models by using $\vartheta _u^p$ ,$\vartheta _a^p$ and $\vartheta _a^n$ instead. Table.\ref{tab_acc_rou_mnist} shows the classification accuracy, where training class priors varies from 0.8 to 1.2 under ground-truth priors. We observer that the proposed model is robust to inaccurate $\theta _u^p$, $\theta _a^p$ and $\theta _a^n$ in mild environment, which prevents the performance degeneration of proposed approach from misspecified mixture proportions.
\subsection{Handling Class Probabilities Shift}
In this section, we investigate class distribution shift in the testing data. Without loss of generality, we conduct the experiments on varying degrees of class distribution shift in the testing phase. Using the  similar setting in section \ref{ritc},  let $\eta_p$, $\eta_n$ and $\eta_a$ be real number around 1, $\pi'_p = \eta_p \pi_p $, $\pi'_n = \eta_n \pi_n $ and $\pi'_a = \eta_a \pi_a $ be the testing class probabilities, and we report experimental results on MNIST by training model using  $\pi_p$, $\pi_n$ and $\pi_a$ but testing model using  $\pi'_p$, $\pi'_n $ and $\pi'_a$ instead. Table.\ref{tab_acc_shift_mnist} reports the performance for handling class distribution shifting in terms of classification accuracy. The results prove that our approach can also overcome the class distribution shift problem in the open environment.

\section{Conclusion}
In this paper, we investigate the problem of learning from positive and unlabeled data with unobserved augmented classes by exploiting augmented classes data. We propose an unbiased risk estimator for positive and unlabeled learning with augmented classes. Besides, we provide a theoretical analysis of estimation error bound, which certainly guarantees the estimator converges to the optimal solution. Experiments demonstrated the effectiveness of proposed methods.
In the future, we will study multi-positive and unlabeled data with unobserved augmented classes, which is common in real-world applications. Besides, an interesting future issue is to investigate the advanced method for PUAC without augmented classes data.

\bibliographystyle{plainnat}
\bibliography{nips}

\end{document}